\documentclass[11pt]{article}

\usepackage[final]{acl}

\usepackage{times}
\usepackage{latexsym}
\usepackage{stmaryrd}

\usepackage[T1]{fontenc}

\usepackage[utf8]{inputenc}

\usepackage{microtype}

\usepackage{inconsolata}
\usepackage{hyperref}

\usepackage{amsmath,amssymb,amsthm,amsfonts}
\usepackage{graphicx}
\usepackage{booktabs}
\usepackage{algorithm}
\usepackage{algpseudocode}
\usepackage{enumitem}
\usepackage{comment}
\usepackage[misc]{ifsym} 
\usepackage[inkscapelatex=false]{svg}
\usepackage{url}
\usepackage{textcomp}
\usepackage{xcolor}
\usepackage{multirow}
\usepackage{tabularx}

\newtheorem{definition}{Definition}

\title{ATOM: AdapTive and OptiMized dynamic temporal knowledge graph construction using LLMs}

\author{\textbf{Yassir LAIRGI}$^{1,2}$, \textbf{Ludovic MONCLA}$^1$, \textbf{Khalid BENABDESLEM}$^1$, \\
  \textbf{Rémy CAZABET}$^1$, \and \textbf{Pierre CLÉAU}$^2$ \\
  $^1$INSA Lyon, CNRS, UCBL, LIRIS, UMR5205, 69621 Villeurbanne,
France \\
  $^2$GAUC, Lyon, France \\
  \texttt{\{ludovic.moncla, khalid.benabdeslem, remy.cazabet\}@liris.cnrs.fr} \\
  \texttt{\{yassir.lairgi, pierre.cleau\}@auvalie.com}}
\begin{document}
\maketitle
\begin{abstract}

In today’s rapidly expanding data landscape, knowledge extraction from unstructured text is vital for real-time analytics, temporal inference, and dynamic memory frameworks. However, traditional static knowledge graph (KG) construction often overlooks the dynamic and time-sensitive nature of real-world data, limiting adaptability to continuous changes. Moreover, recent zero- or few-shot approaches that avoid domain-specific fine-tuning or reliance on prebuilt ontologies often suffer from instability across multiple runs, as well as incomplete coverage of key facts. To address these challenges, we introduce \texttt{ATOM}\footnote{The code, prompts, and dataset are available at \url{https://github.com/AuvaLab/itext2kg}. \texttt{ATOM} is available as an open-source Python library.} (AdapTive and OptiMized), a few-shot and scalable approach that builds and continuously updates Temporal Knowledge Graphs (TKGs) from unstructured texts. \texttt{ATOM} splits input documents into minimal, self-contained “atomic” facts, improving extraction exhaustivity and stability. Then, it constructs atomic TKGs from these facts, employing a dual-time modeling that distinguishes between when information is observed and when it is valid. The resulting atomic TKGs are subsequently merged in parallel. Empirical evaluations demonstrate that \texttt{ATOM} achieves $\sim 18\%$ higher exhaustivity, $\sim 33\%$ better stability, and over 90\% latency reduction compared to baseline methods, demonstrating a strong scalability potential for dynamic TKG construction.

\end{abstract}
\section{Introduction}
Unstructured data is expanding at an unprecedented rate \citep{dresp2019occam}, and given that the majority of big data is inherently unstructured \citep{trugenberger2015scientific}, there is an urgent need for robust information extraction and data modeling techniques to unlock its potential and derive insights across a broad spectrum of applications \citep{cetera2022potential}. A prominent model for converting this unstructured data into structured, actionable knowledge is the Knowledge Graph (KG) \citep{zhong2023comprehensive}.

KG construction involves identifying entities, relationships, and attributes from diverse data sources to create structured knowledge representations. Traditionally, many approaches have focused on static KGs, which provide snapshots of knowledge without incorporating temporal dynamics. However, as real-world phenomena are inherently dynamic, static KGs, rarely or never updated, struggle to remain relevant and accurate \citep{jiang2023evolution}. In contrast, Temporal Knowledge Graphs (TKGs) integrate time dimensions by associating timestamps or time intervals with facts (e.g., (Einstein, was awarded, the Nobel Prize, in 1921)), making them particularly well-suited for analyzing changes, trends, and enabling temporal reasoning.

GraphRAG \citep{edge2024local} and agent-based architectures \citep{xi2023rise} have demonstrated the potential of TKGs in retrieving and modeling dynamic information \citep{wu2024retrieval}. Additionally, TKGs have been effectively used to model the memory of agents within agentic systems \citep{anokhin2024arigraph}, highlighting their role in capturing the evolving nature of knowledge for adaptive and responsive systems.

Despite this potential, deploying Dynamic TKGs in production imposes latency constraints. As new information arrives continuously, systems must update the graph in real-time. In such contexts, sequential approaches often become prohibitive bottlenecks, creating a critical limitation for scalable, real-time analytics \citep{bian2025llm}.

Traditional methods for KG construction, often reliant on entity recognition and relation extraction, face several limitations. They typically depend on predefined ontologies and supervised learning techniques that require extensive human annotation \citep{8999622}. Recent advances in Large Language Models (LLMs) \citep{jin2023large} and zero- or few-shot techniques \citep{zhang2024attackg+, carta2023iterative,hu2023LLM} have paved the way for more flexible KG construction approaches that reduce dependency on extensive training datasets.

Despite these advances, current zero- or few-shot methods for KG construction often suffer from several limitations. They can be non-exhaustive, omitting key relationships, and prone to instability, where multiple construction runs on the same text yield different results. Moreover, many of these approaches overlook the temporal dimension of the input data and struggle to adapt to real-life scenarios with dynamic, evolving data, leading to false positives and a lack of scalability \citep{cai2024survey}.

In this paper, we propose \texttt{ATOM} (AdapTive and OptiMized), a few-shot and scalable dynamic TKG construction approach from unstructured text, ensuring stability and exhaustivity. \texttt{ATOM} introduces a strategy that decomposes unstructured text into atomic facts. Rather than processing these atomic facts sequentially, \texttt{\texttt{ATOM}} proposes an architecture with parallel 5-tuple extraction, followed by a parallel atomic merging mechanism. In the remainder of the paper, we present related work in Section~\ref{related-work}, our proposed approach, \texttt{ATOM}, in Section~\ref{proposed-appraoch}, experimental evaluation in Section~\ref{experiments}, a conclusion in Section~\ref{conclusion}, and limitations in Section~\ref{discussion}.

\section{Related work}
\label{related-work}
Current zero- and few-shot approaches to KG construction, such as AttacKG+ \citep{zhang2024attackg+}, iterative LLM prompting pipelines \citep{carta2023iterative}, LLM-Tikg \citep{hu2023LLM}, LLM Builder \footnote{\url{https://llm-graph-builder.neo4jlabs.com/}}, and LLM Graph Transformer\footnote{\url{https://python.langchain.com/docs/how_to/graph_constructing/}} aim to build KGs without requiring task-specific training. However, these methods suffer from inconsistencies such as unresolved entities and relations.

That is why iText2KG \cite{lairgi2024itext2kg} introduces an incremental, zero-shot architecture that constructs KGs iteratively by comparing newly extracted entities and relations with existing ones using embeddings and cosine similarity, achieving performance gains over some state-of-the-art LLM-based methods. However, it produces non-exhaustive and non-stable KGs due to the stochastic nature of LLMs \cite{atil2024non}. Moreover, it fails to incorporate the temporal dimension, and scalability remains a significant challenge when applying it to real-world scenarios due to its incremental nature.

Graphiti \citep{rasmussen2025zep} proposed a dynamic TKG construction approach for agents' memory with an exclusively LLM-based entity/relation and temporal resolution framework. A key limitation is that it relies solely on prompting the LLM across all its modules, making it heavily dependent on LLM calls. The system prompts the LLM with all previous entities for entity resolution, which becomes impractical as the graph scales to millions of nodes. Similarly, time conflicts are resolved exclusively through LLM calls, resulting in high computational costs and scalability challenges when applied to large-scale datasets. Moreover, they do not handle the exhaustivity and stability of the constructed TKGs.

AriGraph \citep{anokhin2024arigraph} integrated semantic and episodic memories to support reasoning, planning, and decision-making in LLM agents. However, their entity resolution method leads to semantic drift in the temporal KG, where, for example, a reference to “Apple” might ambiguously denote either the company or the fruit. Furthermore, scalability becomes problematic as the volume of unstructured data increases.

Despite these advances, current zero- and few-shot TKG construction methods face three key limitations: (1) they struggle to maintain exhaustive fact coverage when processing longer texts, (2) they often produce unstable TKGs across multiple runs, and (3) they lack scalable architectures for dynamic temporal updates. To address these challenges, we propose \texttt{ATOM}, a framework that combines atomic fact decomposition for exhaustive and stable extraction and parallel merging for scalability.

\section{Proposed approach: \texttt{ATOM}}
\label{proposed-appraoch}
In this section, we first present some notations and definitions used throughout the paper and then introduce the formulation of our proposed framework.
\begin{figure*}[!t]
  \centering
  \includegraphics[width=\textwidth]{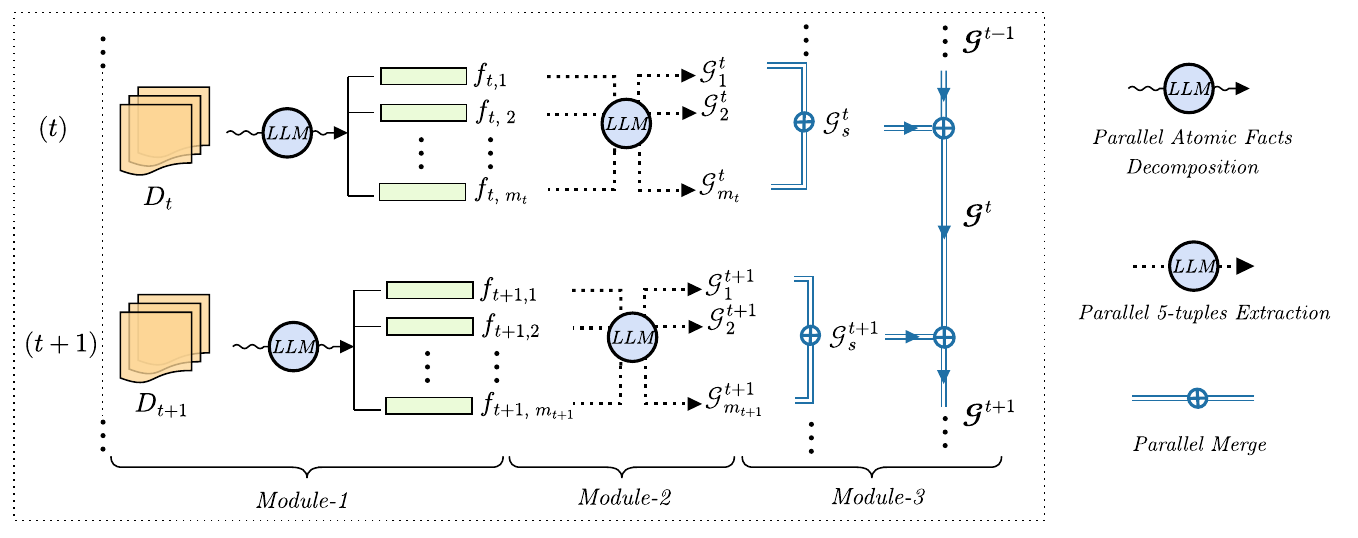}
  \caption{
    \label{fig:atomarchitecture}
    ATOM’s architecture, running in
    parallel, ensuring scalability, speed, and continuous updates. Unstructured texts observed at time $t$ are denoted by $D_t$, the $i$-th temporal atomic fact observed at time $t$ is denoted by $f_{t,i}$, the $i$-th atomic TKG observed at time $t$ is denoted by $\mathcal{G}^t_i$, the TKG snapshot observed at time $t$ is denoted by $\mathcal{G}_{s}^t$, and the updated TKG at time $t$ is denoted by $\mathcal{G}^t$.
  }
\end{figure*}

\subsection{Problem statement}
\texttt{ATOM} incorporates dual-time modeling, differentiating between when facts are observed and the temporal information conveyed by the facts themselves, which is characterized by a validity period. This approach better reflects real-world data \citep{rasmussen2025zep, chekol2018towards, meijer2022bi}. This separation enhances TKG dynamism and proper inference of relative times by providing the observation time as context to the LLM (eg, 'month ago').

\begin{definition}[Dynamic-Temporal KG with Dual-Time Modeling]
\label{def:twoTimeDTKG}
Let $\mathcal{T}_\text{obs}$ be an ordered set of \emph{observation timestamps} at which the KG is updated, and let $\mathcal{T}_{\text{start}}$, $\mathcal{T}_{\text{end}}$ be sets of timestamps used to label inherent validity period of facts defined by their start and end times, respectively. For each observation time $t \in \mathcal{T}_\text{obs}$, a \emph{TKG} snapshot is defined as:

\begin{equation}
    \mathcal{G}_{s}^t  
  \;=\;  
  \Bigl(
    \mathcal{E}^t,\;  
    \mathcal{R}^t,\;  
    \mathcal{T}_{\text{start}}^t,\;
    \mathcal{T}_{\text{end}}^t,\;  
    \mathcal{F}^t
  \Bigr)
\end{equation}

where:
\begin{itemize}
    \item $\mathcal{E}^t$ is the set of entities known at \emph{observation time} $t$,
    \item $\mathcal{R}^t$ is the set of relations known at \emph{observation time} $t$,
    \item $\mathcal{T}_{\text{start}}^t \subseteq \mathcal{T}_{\text{start}}$ is the set of \emph{validity start times} referenced by facts in this snapshot,
    \item $\mathcal{T}_{\text{end}}^t \subseteq \mathcal{T}_{\text{end}}$ is the set of \emph{validity end times} referenced by facts in this snapshot,
    \item $\mathcal{F}^t \;\subseteq\; \mathcal{E}^t \times \mathcal{R}^t \times \mathcal{E}^t \times {\mathcal{T}_{\text{start}}^t} \times {\mathcal{T}_{\text{end}}^t}$ is the set of temporal facts (5-tuples) observed in the snapshot at \emph{observation time} $t$.
\end{itemize}

A fact in this snapshot is a 5-tuple (quintuple ) \((e_s, r_p, e_o, t_{\text{start}}, t_{\text{end}})\) indicating the relation \(r_p \in \mathcal{R}^t\) holds between the subject entity \(e_s \in \mathcal{E}^t\) and the object entity \(e_o \in \mathcal{E}^t\). Technically, $t_{\text{start}}$ and $t_{\text{end}}$ are chosen to be lists to aggregate start and end validity timestamps to track the history of the same fact. Validity start and end timestamps can be unknown, in which case their respective lists are empty (denoted as $[.]$).

A \emph{Dynamic Temporal Knowledge Graph (DTKG)}, updated at t, is defined as the parallel pairwise merge of these TKG snapshots via the merge operator $\oplus$ (as described later in Section~\ref{sec:module3}).
\begin{equation}
  \mathcal{G}^t = \bigoplus_{t' \in \mathcal{T}_{obs} = \{..., t-1, t\}} \mathcal{G}_{s}^{t'} =  \mathcal{G}^{t-1} \oplus \mathcal{G}_{s}^{t}
\end{equation}
\end{definition}

\begin{definition}[Temporal Atomic Fact with Dual-Time Modeling]
\label{def:atomic-fact}
Let $\mathcal{T}_{\mathrm{obs}}$ be a set of \emph{observation timestamps} at which new data is ingested, and let $\mathcal{T}_{\mathrm{start}}$ and $\mathcal{T}_{\mathrm{end}}$ be the sets of validity periods mentioned within the data. For each observation time $t \in \mathcal{T}_{\mathrm{obs}}$, let $D_{t}$ be an unstructured text that becomes available at $t$. A \emph{temporal atomic fact} 
\(
  f_{t,i}
\)
is a short, self-contained snippet derived from $D_{t}$ that conveys exactly one piece of information, extracted by the LLM. Depending on the content of the snippet, an atomic fact may or may not explicitly contain a validity period. Formally,
\begin{equation}
    \mathsf{ExtractAFacts_{LLM}}(D_{t}) = \{f_{t,1}, \dots, f_{t,m_{t}}\}
\end{equation}
An example is provided in Section~\ref{sec:example-afact-decomp} in the Appendices. In what follows, the term atomic fact is used for conciseness. 
\end{definition}

\begin{definition}[Atomic Temporal KG]\label{def:atomic-kg}
Given an atomic fact $f_{t,i}$ observed at time $t$, its \emph{atomic temporal KG} $\mathcal{G}^{t}_{i}$ is the set of 5-tuples extracted by the LLM:
\begin{align}
    \mathcal{G}^{t}_{i} &= \mathsf{ExtractQuintuples_{LLM}}(f_{t,i}) \\
    &\subseteq \mathcal{P}\bigl(\mathcal{E}^t \times \mathcal{R}^t \times \mathcal{E}^t \times {\mathcal{T}_{\text{start}}^t} \times {\mathcal{T}_{\text{end}}^t} \bigr)
\nonumber
\end{align}
Concretely, $\mathcal{G}^{t}_{i}$ is the set of 5-tuples $(e_s, r_p, e_o, t_{\text{start}}, t_{\text{end}})$ derived from a single atomic fact $f_{t,i}$ at observation time $t$.
\end{definition}

Given the definitions~\ref{def:twoTimeDTKG} and~\ref{def:atomic-kg}, the DTKG, updated at $t$:

\begin{equation}
  \mathcal{G}^t = \bigoplus_{t' \in \mathcal{T}_{obs}} \mathcal{G}_{s}^{t'} = \bigoplus_{t' \in \mathcal{T}_{obs}} (\bigoplus_{i \in \llbracket 1, m_{t'}\rrbracket} \mathcal{G}_{i}^{t'})
\end{equation}

\begin{figure*}[h]
\label{fig:example_of_atom_pipeline}
  \centering
  \includegraphics[width=\textwidth]{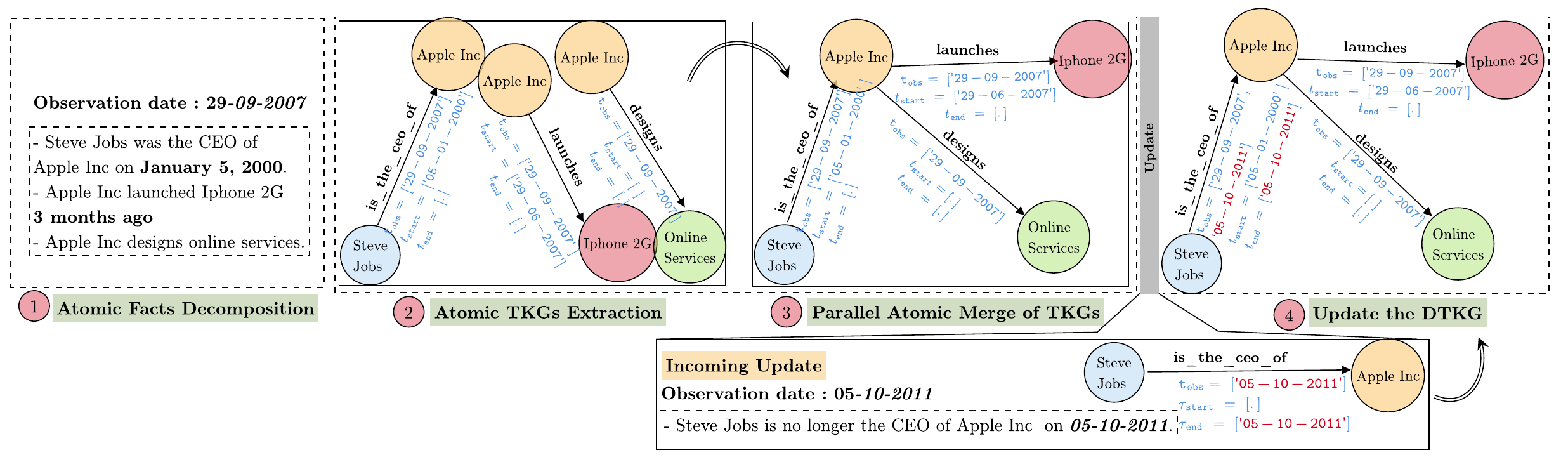}
  \caption{
    \label{fig:example_atom}
     Example overview of \texttt{ATOM}'s pipeline. It begins with atomic fact decomposition, followed by the extraction of atomic TKGs from these facts, which are then merged in parallel. When an incoming update arrives, \texttt{ATOM} handles the temporal resolution by transforming the end action into the affirmative part while modifying only the $t_{\text{end}}$, then merges the resulting atomic TKG with the existing DTKG.
  }
\end{figure*}

Figure~\ref{fig:example_atom} illustrates a detailed example of \texttt{ATOM}'s pipeline. 

\subsection{\texttt{ATOM}'s Framework}
Given a continuous stream of unstructured texts, our goal is to construct and maintain a consistent and dynamic TKG, ensuring for each $t \in \mathcal{T}_{obs}$:
\begin{enumerate}[label= (C\arabic*)]
    \item \label{c1} Exhaustivity: the constructed TKG snapshot $\mathcal{G}_{s}^t$ ideally captures every 5-tuple that is semantically present in $D_t$. 
    \item \label{c2} Stability across multiple runs: when the identical 5-tuple extraction prompt is executed repeatedly on the same input text using the same LLM, the resulting TKG snapshots should be nearly identical. 
\end{enumerate}
In the following,  we detail the different modules of  \texttt{ATOM}'s architecture (Figure~\ref{fig:atomarchitecture}).

\subsubsection{Module-1: Atomic fact decomposition}
\label{step1}
\texttt{ATOM} does not construct TKGs directly from raw input documents but first decomposes them into atomic facts (Figure~\ref{fig:atomarchitecture}). This decomposition addresses a fundamental limitation of LLMs: the "forgetting effect" where models prioritize salient information in longer contexts while omitting key relationships, leading to incomplete knowledge extraction \cite{liu2023lost}. Following \citep{hosseini2024scalable, chen2023dense, raina2024question}, \texttt{ATOM} uses LLM-based prompting for decomposition with an optimal chunk size to maintain high exhaustivity (determined experimentally in Section~\ref{sec:evaluating_exhaustivity}). However, while prior work focused on information retrieval applications, \texttt{ATOM} applies atomic decomposition specifically for TKG construction. This strategy addresses both conditions: it enhances exhaustivity~\ref{c1} by preventing information loss that occurs when LLMs process complex, multi-fact paragraphs, and enhances stability~\ref{c2} by providing clear, unambiguous contexts that reduce output variance across multiple runs. Each atomic fact is related to an observation time, and it is necessary to encapsulate the relative validity period presented in the context. The primary computational challenge of this approach is scale: a single document can yield hundreds or thousands of atomic facts. Sequential processing of each fact for 5-tuple extraction, followed by entity/relation and temporal resolution, becomes time-consuming. To address this challenge, \texttt{ATOM} employs a parallel architecture for both extraction and merging phases, as detailed in the subsequent modules.

\subsubsection{Module-2: Atomic TKGs construction}
\label{sec:temporal_reso}
5-tuples are extracted from each atomic fact in parallel using an LLM, producing atomic TKGs $\mathcal{G}^t_i$ while embedding nodes and relations following \citep{lairgi2024itext2kg}. To facilitate temporal resolution in Module-3, \texttt{ATOM} preprocesses 5-tuples during their extraction. It prevents separate quintuples describing the same temporal fact from coexisting in the same TKG such as (John\_Doe, is\_ceo, X, [01-01-2025], [.]) and (John\_Doe, is\_no\_longer\_ceo, X, [01-01-2026], [.]), which should be resolved into (John\_Doe, is\_ceo, X, [01-01-2025], [01-01-2026]). During the extraction, few-shot examples are provided as context to the LLM to transform end validity facts into affirmative counterparts while modifying only the $t_{\text{end}}$ time. For instance, the statement John Doe is no longer the CEO of X on 01-01-2026 is converted into the 5-tuple (John\_Doe, is\_ceo, X, [.], [01-01-2026]), ensuring direct matching with the corresponding validity start time 5-tuple during the merge. For relative temporal expressions (e.g., 'a month ago'), the observation time is provided as context, enabling the LLM to infer the validity period.

\subsubsection{Module-3: Parallel atomic merge of TKGs and DTKG update}
\label{sec:module3}
\texttt{ATOM} then employs the binary merge algorithm (Algorithm~\ref{alg:binary_atomic_merge} in the Appendices) to merge pairs of atomic TKGs. The algorithm proceeds in three phases: first, entity resolution searches for exact matches between $\mathcal{G}^t_i$ and $\mathcal{G}^t_{i+1}$ based on name and label. When no exact match exists, cosine similarity is computed, merging entities if similarity exceeds $\theta_E$. Second, relation resolution merges relation names regardless of endpoints and timestamps (e.g., owns $\longleftrightarrow$ possesses $\longleftrightarrow$ has) using threshold $\theta_R$. Third, temporal resolution merges observation and validity time sets for relations with similar $(e_s, r_p, e_o)$, detecting and aligning end-action facts with their corresponding beginning facts. Unlike Graphiti, \texttt{ATOM} avoids LLM calls during merging, improving scalability and preventing context overflow as the graph expands. The preprocessing of end-actions during extraction enables this LLM-independent merging approach. Subsequently, the binary merge function is extended to handle the entire set of atomic TKGs through iterative pairwise merging in parallel until a single consolidated TKG is obtained (Algorithm~\ref{alg:parallel_atomic_merge} in the Appendices). Atomic TKGs are organized into pairs, with each pair merged in parallel. If the number of TKGs is odd, the remaining TKG carries forward to the next iteration. This process continues iteratively, reducing the number of TKGs at each step, until convergence to a single merged TKG. This parallel strategy scales with the number of available threads and addresses the computational challenge from Module-1, enabling \texttt{ATOM} to maintain low latency while preserving the exhaustivity and stability benefits of atomic decomposition. After the merge of all atomic TKGs, the snapshot $\mathcal{G}^t_{s}$ is obtained, and it is merged with the previous DTKG $\mathcal{G}^{t-1}$ using Algorithm~\ref{alg:binary_atomic_merge} to yield the DTKG updated at $t$, $\mathcal{G}^{t}$.

\section{Experiments} 
\label{experiments}
Our evaluation addresses the following research questions:

\begin{enumerate}
\item[\textbf{RQ1:}] How does exhaustivity deteriorate as the LLM context increases, and what degree of information loss could occur?
\item[\textbf{RQ2:}] How does \texttt{ATOM}'s atomic fact decomposition enhance stability, exhaustivity, and improve the quality of the 5-tuples?
\item[\textbf{RQ3:}] How does \texttt{ATOM} scale with the number of atomic facts provided as input, and what is its time complexity compared to baseline methods?
\item[\textbf{RQ4:}] How does \texttt{ATOM} perform on DTKG construction consistency compared to baseline methods?
\end{enumerate}

\subsection{Metrics}
\label{sec:metrics}

To assess~\ref{c1}, the performance is evaluated at two levels: factual and factual-temporal. Given a gold-standard KG, factual true positives ($\text{TP}_f$) denote correctly extracted triplets, false negatives ($\text{FN}_f$) refer to triplets that exist in the source but are missing from the extracted KG (omission), and false positives ($\text{FP}_f$) represent unsupported triplets (hallucination). These definitions are extended to the factual-temporal level for 5-tuples $(e_s, r_p, e_o, t_{\text{start}}, t_{\text{end}})$. A 5-tuple is considered valid only if it exhibits correctness at both the factual and temporal levels. Hence, the factual-temporal true positives $\text{TP}_{f,t}$ are 5-tuples whose $t_{\text{start}}$ and $t_{\text{end}}$ are both correct. Consequently, $\text{TP}_{f,t} \subseteq \text{TP}_f$.
    
\paragraph{Exhaustivity (RQ1 \& RQ2).}
The exhaustivity or recall is reported at two levels of strictness. The exhaustivity of the factual component:
\begin{equation}
R_f = \frac{|\text{TP}_f|}{|\text{TP}_f| + |\text{FN}_f|}
\end{equation}

The exhaustivity of the factual-temporal component $R_{f,t}$ is computed analogously with $|\text{TP}_{f,t}|$ in the numerator. It measures the proportion of the total gold standard 5-tuples that are both factually and temporally correct.

\paragraph{5-tuples quality (RQ2).}
It is assessed through Recall, Precision, and $F_1$-score. The precision of the factual component:
\begin{equation}
P_f = \frac{|\text{TP}_f|}{|\text{TP}_f| + |\text{FP}_f|}
\end{equation}

The precision of the factual-temporal component $P_{f,t}$ is computed similarly to $P_f$ with $|\text{TP}_{f,t}|$ in the numerator. It measures the proportion of all extracted 5-tuples that are both factually and temporally correct. The $F_{1,f}$ and $F_{1,f,t}$ scores are computed as the harmonic means of their respective precision and recall metrics.

\paragraph{Stability (RQ2).}

It is measured using the Jaccard similarity between the 5-tuple sets obtained across independent runs. Let $\boldsymbol{s}^{(1)}$ denote the set of 5-tuples obtained during the baseline run (RUN 1) and let $\boldsymbol{s}^{(r)}$ denote the set of 5-tuples obtained at repetition $r$. Formally, the stability score $S_r$ is computed as:

\begin{equation}
S_r = J\bigl(\boldsymbol{s}^{(1)},\,\boldsymbol{s}^{(r)}\bigr) 
= \frac{\bigl\lvert \boldsymbol{s}^{(1)} \cap \boldsymbol{s}^{(r)} \bigr\rvert}{\bigl\lvert \boldsymbol{s}^{(1)} \cup \boldsymbol{s}^{(r)} \bigr\rvert}
\end{equation}

This score is computed for repetitions $r = 2, 3$.

\paragraph{Time complexity (RQ3).} 
Time complexity is evaluated by progressively increasing the number of atomic facts provided as input and measuring the total wall-clock latency required to construct the complete DTKG.

\paragraph{DTKG consistency (RQ4).} 
For entity/relation resolution, the false discovery rate (1-precision) is defined in \citep{lairgi2024itext2kg}. Since they overlook recall and $F_1$-score, we extend the evaluation to include precision ($P$), recall ($R$), and $F_1$-score for both entity resolution (ER) and relation resolution (RR), denoted as $\text{Metric}_{ER}$ and $\text{Metric}_{RR}$, respectively. For temporal resolution, a qualitative comparison is provided.

\subsection{Datasets and baseline methods}

\label{sec:data_baselines}
Identifying a suitable and publicly available dataset for evaluating temporal extraction remains a significant challenge due to the task's specific structural requirements. While DocRed \citep{yao2019docred} is a standard benchmark for relation extraction, it is largely unsuitable for temporal analysis owing to documented labeling inconsistencies \citep{tan2022revisiting}. Although TempDocRed \citep{zhu2025temporal} introduces temporal layers, its scope remains limited, focusing exclusively on named entities and event start dates. Furthermore, the CS-GS and Music-GS datasets \citep{kabal2024enhancing}, lack the necessary temporal dimensions. Consequently, this study adopts the NYT News dynamic and temporal dataset \citep{AryanSingh2023NYT}, which provides extensive temporal coverage through two million lead paragraphs spanning from 2000 to the present.
News articles provide diverse temporal dynamics, including events with specific start and end times, evolving situations, and relative temporal expressions that are representative of many real-world scenarios requiring TKG construction. From this dataset, the 2020-COVID-NYT subset comprising 1,076 articles that focus on COVID-19 dynamics during 2020 is extracted. This subset is enriched with human-verified atomic facts and 5-tuples. Publication dates are used as observation dates. Comprehensive details regarding the annotation guidelines, observation time modeling, and dataset statistics are provided in the Appendices (Sections~\ref{app:annotation_guidelines}, \ref{sec:observation-time-modeling}, and Table~\ref{tab:dataset_statistics}). To the best of our knowledge, an approach for resolving duplicate entities and relations while maintaining KG consistency among the SOTA methods for zero- and few-shot KG construction is supported only by iText2KG and Graphiti. The temporal aspect is handled by Graphiti only. Hence, it is the primary comparator of \texttt{ATOM}. In all experiments, the temperature is set to 0 to prefer deterministic outputs and minimize stochasticity. \texttt{text-embedding-large-3}\footnote{\url{https://platform.openai.com/docs/models/text-embedding-3-large}} is used for embeddings. $\theta_E = 0.8$ and $\theta_R=0.7$ are estimated following \cite{lairgi2024itext2kg} (details are in Section~\ref{sec:thresholds-estimation} in the Appendices).

\subsection{Exhaustivity deterioration in longer contexts}
\label{sec:evaluating_exhaustivity}

\begin{figure}[t]
\centering
\includegraphics[width=\columnwidth]{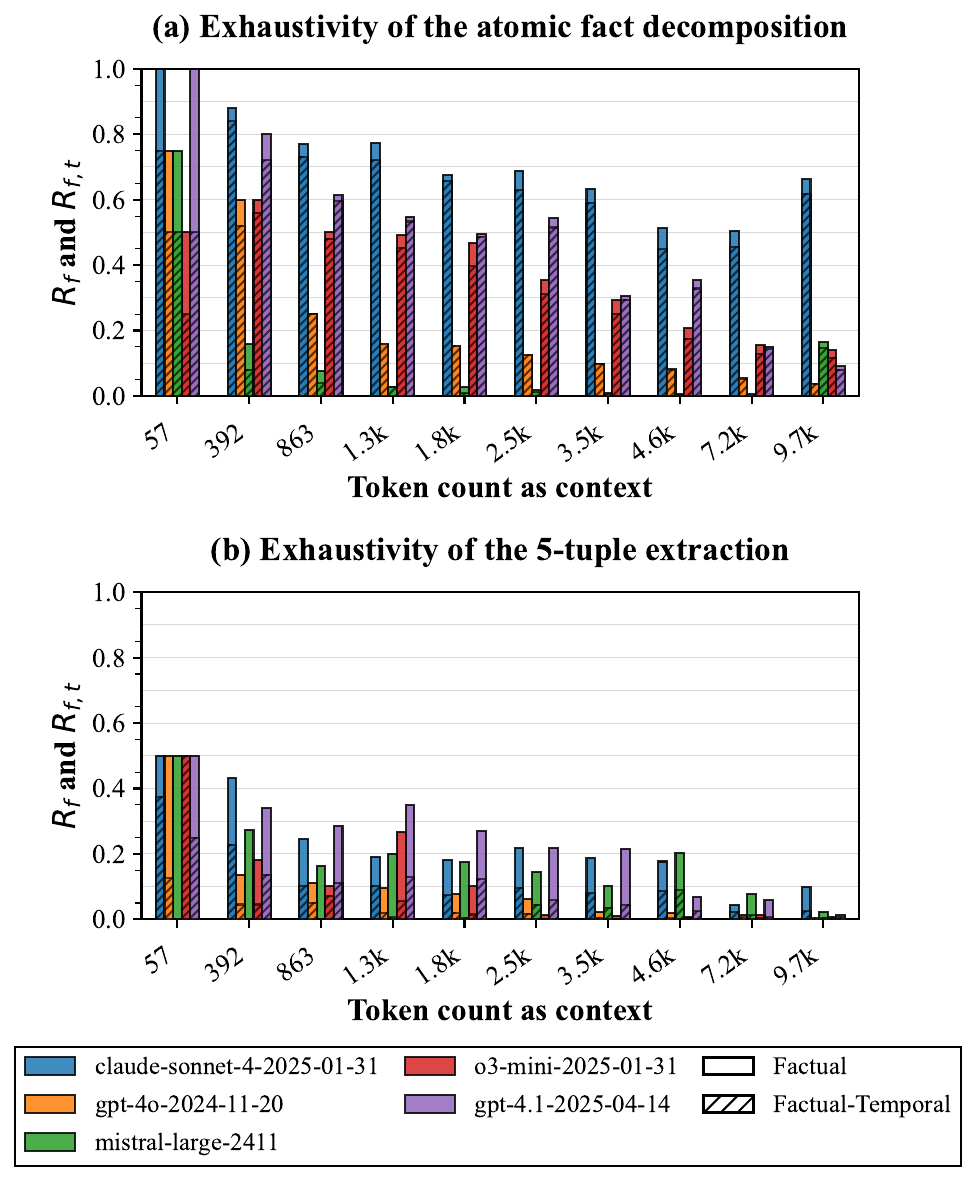}
\caption{\label{fig:exhaustivity} Exhaustivity vs. token count as context for (a) the atomic fact decomposition (b) the 5-tuple extraction. The x-axis displays the context size (lead paragraphs) only; total input size includes additional system prompt tokens ($\sim$1.4k for (a) and $\sim$1.7k for (b)).}
\end{figure}

Exhaustivity is evaluated by iteratively concatenating lead paragraphs (increasing context size) and testing five SOTA LLMs \texttt{claude-sonnet-4-2025-01-31}\footnote{\url{https://docs.claude.com/en/docs/about-claude/models/overview}}, \texttt{gpt-4o-2024-11-20}\footnote{\url{https://platform.openai.com/docs/models/gpt-4o-2024-11-20}}, \texttt{mistral-large-2411}\footnote{\url{https://docs.mistral.ai/getting-started/models/models_overview/}}, \texttt{gpt-4.1-2025-04-14}\footnote{\url{https://openai.com/index/gpt-4-1/}}, \texttt{o3-mini-2025-01-31}\footnote{\url{https://openai.com/index/openai-o3-mini/}}. Figure~\ref{fig:exhaustivity} shows a clear "forgetting effect". A decreased factual and factual-temporal exhaustivity as token count increases across all models except \texttt{claude-Sonnet-4-2025-01-31}, which maintains the highest exhaustivity for atomic facts but degrades for 5-tuples. This indicates that LLMs prioritize salient facts in longer texts. Moreover, all models show higher exhaustivity for atomic fact decomposition than 5-tuple extraction. Atomic facts require surface-level parsing, while 5-tuples demand deeper semantic understanding of entities, relationship identification, and temporal extraction ($t_{\text{start}}$, $t_{\text{end}}$). This added complexity causes greater information loss. To mitigate information loss in the atomic fact decomposition, we empirically determine the optimal chunk size at $<400$ tokens to keep the exhaustivity $>0.8$. For subsequent evaluations, we use \texttt{claude-sonnet-4-2025-01-31} for atomic fact decomposition based on its superior performance. For 5-tuple extraction, both \texttt{gpt-4.1-2025-04-14} and \texttt{claude-sonnet-4-2025-01-31} perform comparably, hence \texttt{gpt-4.1-2025-04-14} is used due to its lower cost. Section~\ref{sec-quality} evaluates exhaustivity gains from using atomic facts as input.

\subsection{TKG stability}
Rerunning the same prompt multiple times leads to variations in output. To evaluate the stability score, 5-tuples are initially extracted from both atomic facts (denoted as (F) in Table~\ref{tab:stability_reruns_all}) and lead paragraphs individually, establishing a baseline named Run 1. This extraction process is subsequently repeated twice more without altering any parameters. As shown in Table~\ref{tab:stability_reruns_all}, the 5-tuple sets extracted from atomic facts exhibit high Jaccard similarity with the baseline across runs. In contrast, the extraction from lead paragraphs demonstrates significantly lower overlap, revealing high variability. This addresses~\ref{c2} and reflects the effect of atomic facts in maintaining a stable construction of TKGs with a gain of $\sim 33\%$.

\begin{table}[ht]
\centering
\small
\caption{Stability $S_r$ evaluated by rerunning the 5-tuple extraction process multiple times without any modifications, using \texttt{gpt-4.1-2025-04-14} with Run 1 as a baseline. The extraction is performed on (F) atomic facts and (L) lead paragraphs.}
\label{tab:stability_reruns_all}
\begin{tabular}{@{}lccc@{}}
\toprule
\textbf{Dataset} & \textbf{Run 2} & \textbf{Run 3}  \\
\midrule
2020-COVID-NYT (F) & \textbf{0.552 $\pm$ 0.124} & \textbf{0.534 $\pm$ 0.12}  \\
2020-COVID-NYT (L)     & 0.212 $\pm$ 0.181        & 0.218 $\pm$ 0.195          \\ 
\bottomrule
\end{tabular}
\end{table}

\subsection{The exhaustivity and quality of the 5-tuples}
\label{sec-quality}

5-tuples are extracted from both atomic facts (denoted as (F) in Table~\ref{tab:covid-metrics-comparison}) and lead paragraphs (L). As shown in Table~\ref{tab:covid-metrics-comparison}, \texttt{ATOM} significantly outperforms the baseline in exhaustivity, with factual recall $R_f$ increasing by $\sim 31\%$ and factual-temporal recall $R_{f,t}$ by $\sim 18\%$. These gains lead to a substantial improvement in overall extraction quality, resulting in increased factual $F_{1,f}$ and factual-temporal $F_{1,f,t}$ scores. Hence, atomic fact decomposition effectively mitigates information loss, addressing~\ref{c1}. However, this improved exhaustivity presents a trade-off: a decrease in factual precision $P_f$ of $\sim 9\%$. This occurs because the baseline (L) acts conservatively, extracting only the most obvious facts, whereas \texttt{ATOM}'s decomposition forces the LLM to explicitize context. This process can generate 'inferred' atomic facts that are semantically plausible but not strictly present in the gold standard, leading to false positives. Furthermore, a marginal decrease of $\sim 1\%$ in factual-temporal precision $P_{f,t}$ is observed. This is attributed to imperfections during the decomposition phase, where the LLM may fail to assign temporal information to certain atomic facts. Consequently, this error propagates to the 5-tuple extraction. This trade-off is discussed further in Section~\ref{discussion}.

\begin{table}[ht]
\centering
\small
\caption{5-tuple quality metrics. The extraction is performed on (L) lead paragraphs and (F) atomic facts.}
\label{tab:covid-metrics-comparison}
\resizebox{\columnwidth}{!}{%
\begin{tabular}{@{}lccc@{}}
\toprule
\textbf{Metric} & \textbf{2020-COVID-NYT (L)} & \textbf{2020-COVID-NYT (F)} \\
\midrule
$P_f$             & \textbf{0.667 $\pm$ 0.172} & 0.572 $\pm$ 0.128 \\
$R_f$             & 0.405 $\pm$ 0.150          & \textbf{0.720 $\pm$ 0.143} \\
$F_{1,f}$-score     & 0.504                      & \textbf{0.638} \\
\midrule
$P_{f,t}$           & \textbf{0.292 $\pm$ 0.182} & 0.281 $\pm$ 0.130 \\
$R_{f,t}$           & 0.176 $\pm$ 0.123          & \textbf{0.354 $\pm$ 0.165} \\
$F_{1,f,t}$-score & 0.220                      & \textbf{0.313} \\
\bottomrule
\end{tabular}
}
\end{table}

\subsection{\texttt{ATOM}'s time complexity}
\begin{figure}[h!]
\centering
\includegraphics[width=\columnwidth]{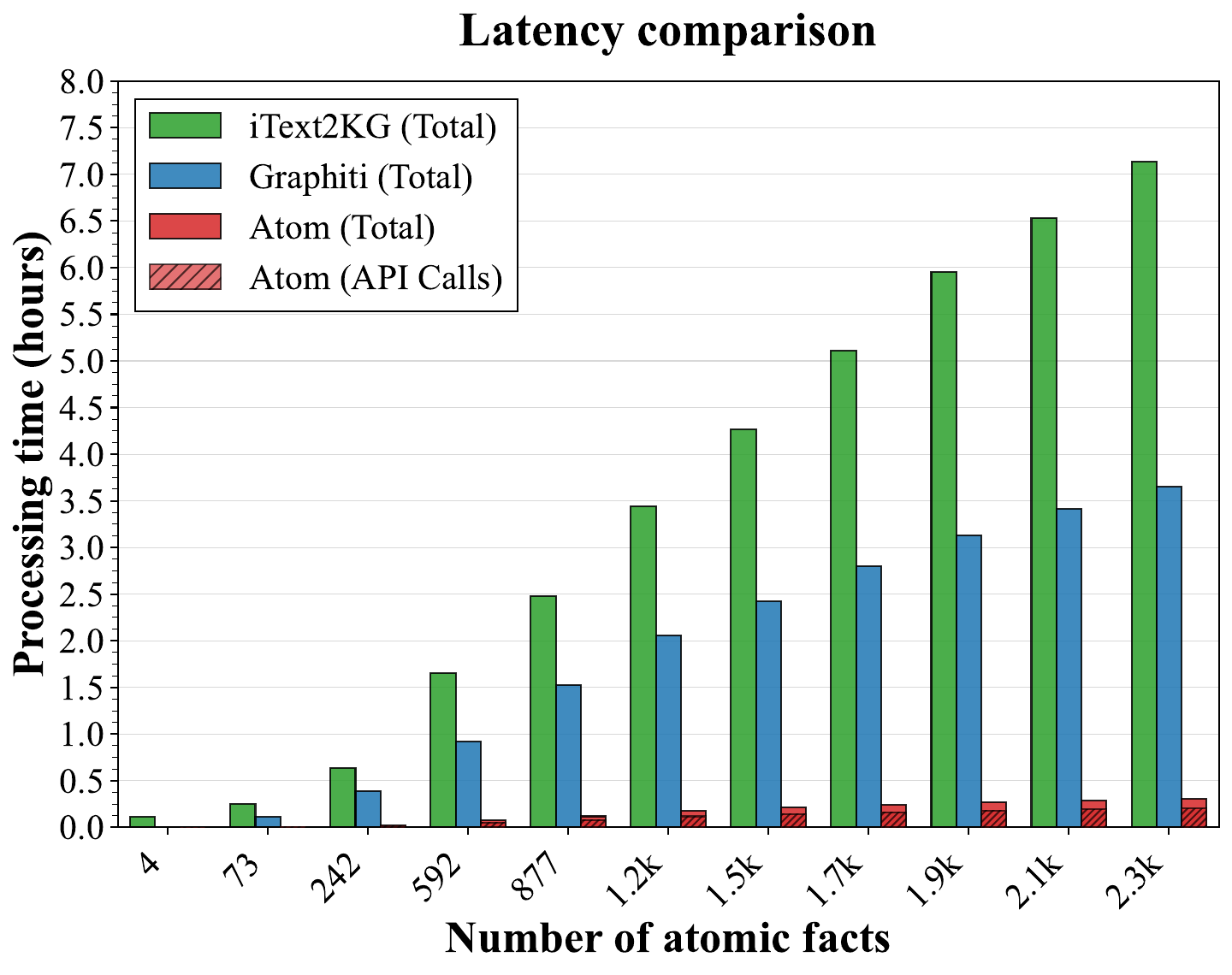}
\caption{\label{fig:latency} Latency comparison of the baseline methods as a function of the number of atomic facts as input.}
\end{figure}

Given the demonstrated benefits of atomic fact decomposition in improving exhaustivity and stability, all subsequent experiments utilize atomic facts as input rather than lead paragraphs. All baseline methods are run using $\texttt{gpt-4.1-2025-04-14}$. \texttt{ATOM} employs 8 threads and a batch size of 40 atomic facts for 5-tuple extraction, which respects OpenAI rate limits. iText2KG and Graphiti separate entity and relation extraction, increasing latency. Graphiti's incremental entity/relation resolution, which relies on the LLM, limits parallel requests and significantly increases latency as the graph expands. Similarly, iText2KG is incremental, restricting parallel requests. Although iText2KG uses a distance metric for resolution, reducing LLM dependency, its separate extraction steps double the number of LLM calls and induce isolated entities that require further LLM iterations. Conversely, \texttt{ATOM}'s architecture facilitates (1) parallel LLM calls, (2) parallel merge of atomic TKGs, (3) LLM-independent merging, and (4) temporal resolution. This design reduces latency by 93.8\% compared to Graphiti and 95.3\% compared to iText2KG (Figure~\ref{fig:latency}). \texttt{ATOM}'s Module-3 accounts for only 13\% of its total latency, with the remainder attributed to API calls, which can be further minimized through either increasing the batch size (by upgrading the API tier) or scaling hardware for local LLM deployment.

\subsection{Consistency of \texttt{ATOM}'s DTKG construction}
Table~\ref{tab:baseline_methods} shows that \texttt{ATOM} and iText2KG demonstrate comparable entity and relation resolution performance, as both employ a distance metric for merging. \texttt{ATOM} shows an improvement over Graphiti, whose incremental, LLM-based entity and relation resolution degrades with graph expansion (increasing context size), which is consistent with findings in Section~\ref{sec:evaluating_exhaustivity}. Beyond entity and relation resolution, temporal resolution reveals more significant differences between \texttt{ATOM} and Graphiti. The examples in Figures~\ref{fig:atomvsgraphiti_death} and~\ref{fig:atomvsgraphiti_protests} in the Appendices illustrate atomic facts observed at different times that refer to the same information but with different validity periods. In these examples, Graphiti creates separate relations for 5-tuples with different validity periods ($t_{\text{start}}$, $t_{\text{end}}$), while \texttt{ATOM} detects these similar relations and extends their validity period history. This temporal resolution serves two functions: tracking events that naturally appear and disappear over time, and matching relations with only $t_{\text{start}}$ or only $t_{\text{end}}$ to complete their validity periods as additional information becomes available. Additionally, Graphiti incorporates validity periods only and does not allow for observation time modeling, treating observation time as $t_{\text{start}}$. This can induce errors. For example, if a news article observed on "January 23, 2020" states "The mysterious respiratory virus spread to at least 10 other countries," Graphiti would set $t_{\text{start}} = 23\text{-}01\text{-}2020$, while the statement does not specify a validity period and the true validity could be weeks or days before its publication. In contrast, \texttt{ATOM} separately models observation time and validity periods, allowing it to recognize atomic facts without a validity period and avoid incorrect temporal assignments (Example in Figure~\ref{fig:atomvsgraphiti} in the Appendices).


\begin{table}[h]
\centering
\small
\caption{Performance on DTKG construction.}
\label{tab:baseline_methods}
\begin{tabular}{@{}lccc@{}}
\toprule
\textbf{Metric} & \textbf{ATOM} & \textbf{Graphiti} & \textbf{iText2KG} \\
\midrule
$P_{ER}$ & 0.994 & 0.967 & 0.974 \\
$R_{ER}$ & 0.993 & 0.952 & 0.980 \\
$F_{1,ER}$-score & 0.994 & 0.959 & 0.977 \\
\addlinespace[1ex]
$P_{RR}$ & 1 & 0.917 & 0.991\\
$R_{RR}$& 1 & 0.888 & 0.988 \\
$F_{1,RR}$-score & 1 & 0.902 & 0.989 \\
%
%
\bottomrule
\end{tabular}
\end{table}


\section{Conclusion}
\label{conclusion}
In this paper, we presented \texttt{ATOM}, a few-shot and scalable approach for constructing and dynamically updating TKGs from unstructured texts. Experimental results indicate that \texttt{ATOM}'s atomic fact decomposition effectively addresses the exhaustivity and stability challenges often observed in LLM-based TKG construction methods. Its parallel architecture accelerates TKG construction and enables scalability for larger unstructured texts. Potential directions for future work include quantitative evaluation of temporal resolution and fine-tuning an LLM specifically for refining atomic fact decomposition. In summary, \texttt{ATOM} enables fast and continuous updates of TKGs.

\section{Limitations}
\label{discussion}
\texttt{ATOM} has some limitations that warrant consideration. First, the atomic fact decomposition can introduce error propagation: the LLM may generate inferred facts not present in the source text, leading to increased hallucination, and may fail to properly assign temporal information to atomic facts, resulting in a decrease in the factual-temporal precision (Section~\ref{sec-quality}). A potential improvement consists of fine-tuning an LLM model specifically for this task. Second, the distance-metric-based merging approach, while scalable and efficient, can occasionally merge semantically distinct named entities that exhibit high similarity (e.g., \texttt{gpt-5:model} and \texttt{gpt-3.5:model}). A supervised entity/relation resolution classifier trained on labeled entity pairs could replace the threshold-based approach. Third, \texttt{ATOM} entails higher costs than direct extraction (Table~\ref{tab:cost_estimation_detailed} in the Appendices) due to the added decomposition step and the increased output tokens required to maintain high exhaustivity. This cost could be mitigated by using or fine-tuning open-weight models specifically for extracting 5-tuples from atomic facts.

\section{Ethics statement}
We utilized AI assistants exclusively for grammatical corrections and text polishing. All scientific claims, experimental designs, and original ideas are solely those of the authors.

\bibliography{custom}

@article{cai2024survey,
  title={A Survey on Temporal Knowledge Graph: Representation Learning and Applications},
  author={Cai, Li and Mao, Xin and Zhou, Yuhao and Long, Zhaoguang and Wu, Changxu and Lan, Man},
  journal={arXiv preprint arXiv:2403.04782},
  year={2024}
}

@article{jiang2023evolution,
  title={On the Evolution of Knowledge Graphs: A Survey and Perspective},
  author={Jiang, Xuhui and Xu, Chengjin and Shen, Yinghan and Sun, Xun and Tang, Lumingyuan and Wang, Saizhuo and Chen, Zhongwu and Wang, Yuanzhuo and Guo, Jian},
  journal={arXiv preprint arXiv:2310.04835},
  year={2023}
}

@inproceedings{lairgi2024itext2kg,
  title={itext2kg: Incremental knowledge graphs construction using large language models},
  author={Lairgi, Yassir and Moncla, Ludovic and Cazabet, R{\'e}my and Benabdeslem, Khalid and Cl{\'e}au, Pierre},
  booktitle={International Conference on Web Information Systems Engineering},
  pages={214--229},
  year={2024},
  organization={Springer}
}

@article{dresp2019occam,
  title={Occam’s Razor for Big Data? On detecting quality in large unstructured datasets},
  author={Dresp-Langley, Birgitta and Ekseth, Ole Kristian and Fesl, Jan and Gohshi, Seiichi and Kurz, Marc and Sehring, Hans-Werner},
  journal={Applied Sciences},
  volume={9},
  number={15},
  pages={3065},
  year={2019},
  publisher={MDPI}
}

@article{trugenberger2015scientific,
  title={Scientific Discovery by Machine Intelligence: A New Avenue for Drug Research},
  author={Trugenberger, Carlo A},
  journal={arXiv preprint arXiv:1506.07116},
  year={2015}
}

@article{cetera2022potential,
  title={Potential for the use of large unstructured data resources by public innovation support institutions},
  author={Cetera, Wies{\l}aw and Gogo{\l}ek, W{\l}odzimierz and {\.Z}o{\l}nierski, Aleksander and Jaruga, Dariusz},
  journal={Journal of Big Data},
  volume={9},
  number={1},
  pages={46},
  year={2022},
  publisher={Springer}
}

@article{zhong2023comprehensive,
  title={A comprehensive survey on automatic knowledge graph construction},
  author={Zhong, Lingfeng and Wu, Jia and Li, Qian and Peng, Hao and Wu, Xindong},
  journal={ACM Computing Surveys},
  volume={56},
  number={4},
  pages={1--62},
  year={2023},
  publisher={ACM New York, NY}
}

@ARTICLE{8999622,
  author={Al-Moslmi, Tareq and Gallofré Ocaña, Marc and L. Opdahl, Andreas and Veres, Csaba},
  journal={IEEE Access}, 
  title={Named Entity Extraction for Knowledge Graphs: A Literature Overview}, 
  year={2020},
  volume={8},
  number={},
  pages={32862-32881},
}

@article{carta2023iterative,
  title={Iterative zero-shot {LLM} prompting for knowledge graph construction},
  author={Carta, Salvatore and Giuliani, Alessandro and Piano, Leonardo and Podda, Alessandro Sebastian and Pompianu, Livio and Tiddia, Sandro Gabriele},
  journal={arXiv preprint arXiv:2307.01128},
  year={2023}
}

@article{hu2023LLM,
title = {LLM-TIKG: Threat intelligence knowledge graph construction utilizing large language model},
journal = {Computers \& Security},
volume = {145},
pages = {103999},
year = {2024},
issn = {0167-4048},
author = {Yuelin Hu and Futai Zou and Jiajia Han and Xin Sun and Yilei Wang},

}

@article{zhang2024attackg+,
  title={{AttacKG+}: Boosting Attack Knowledge Graph Construction with Large Language Models},
  author={Zhang, Yongheng and Du, Tingwen and Ma, Yunshan and Wang, Xiang and Xie, Yi and Yang, Guozheng and Lu, Yuliang and Chang, Ee-Chien},
  journal={arXiv preprint arXiv:2405.04753},
  year={2024}
}

@ARTICLE{jin2023large,
  author={Jin, Bowen and Liu, Gang and Han, Chi and Jiang, Meng and Ji, Heng and Han, Jiawei},
  journal={IEEE Transactions on Knowledge and Data Engineering}, 
  title={Large Language Models on Graphs: A Comprehensive Survey}, 
  year={2024},
  volume={36},
  number={12},
  pages={8622-8642},}

@inproceedings{anokhin2024arigraph,
  title     = {AriGraph: Learning Knowledge Graph World Models with Episodic Memory for LLM Agents},
  author    = {Anokhin, Petr and Semenov, Nikita and Sorokin, Artyom and Evseev, Dmitry and Kravchenko, Andrey and Burtsev, Mikhail and Burnaev, Evgeny},
  booktitle = {Proceedings of the Thirty-Fourth International Joint Conference on
               Artificial Intelligence, {IJCAI-25}},
  publisher = {International Joint Conferences on Artificial Intelligence Organization},
  editor    = {James Kwok},
  pages     = {12--20},
  year      = {2025},
  month     = {8},
  note      = {Main Track},
}

@article{edge2024local,
  title={From local to global: A graph rag approach to query-focused summarization},
  author={Edge, Darren and Trinh, Ha and Cheng, Newman and Bradley, Joshua and Chao, Alex and Mody, Apurva and Truitt, Steven and Larson, Jonathan},
  journal={arXiv preprint arXiv:2404.16130},
  year={2024}
}

@article{xi2023rise,
  title={The rise and potential of large language model based agents: A survey},
  author={Xi, Zhiheng and Chen, Wenxiang and Guo, Xin and He, Wei and Ding, Yiwen and Hong, Boyang and Zhang, Ming and Wang, Junzhe and Jin, Senjie and Zhou, Enyu and others},
  journal={Science China Information Sciences},
  volume={68},
  number={2},
  pages={121101},
  year={2025},
  publisher={Springer}
}

@article{wu2024retrieval,
  title={Retrieval Augmented Generation for Dynamic Graph Modeling},
  author={Wu, Yuxia and Fang, Yuan and Liao, Lizi},
  journal={arXiv preprint arXiv:2408.14523},
  year={2024}
}

@inproceedings{hosseini2024scalable,
    title = "Scalable and Domain-General Abstractive Proposition Segmentation",
    author = {Hosseini, Mohammad Javad  and
      Gao, Yang  and
      Baumg{\"a}rtner, Tim  and
      Fabrikant, Alex  and
      Amplayo, Reinald Kim},
    editor = "Al-Onaizan, Yaser  and
      Bansal, Mohit  and
      Chen, Yun-Nung",
    booktitle = "Findings of the Association for Computational Linguistics: EMNLP 2024",
    month = nov,
    year = "2024",
    address = "Miami, Florida, USA",
    publisher = "Association for Computational Linguistics",
    pages = "8856--8872",

}

@inproceedings{chen2023dense,
    title = "Dense {X} Retrieval: What Retrieval Granularity Should We Use?",
    author = "Chen, Tong  and
      Wang, Hongwei  and
      Chen, Sihao  and
      Yu, Wenhao  and
      Ma, Kaixin  and
      Zhao, Xinran  and
      Zhang, Hongming  and
      Yu, Dong",
    editor = "Al-Onaizan, Yaser  and
      Bansal, Mohit  and
      Chen, Yun-Nung",
    booktitle = "Proceedings of the 2024 Conference on Empirical Methods in Natural Language Processing",
    month = nov,
    year = "2024",
    address = "Miami, Florida, USA",
    publisher = "Association for Computational Linguistics",
    pages = "15159--15177",
}

@inproceedings{raina2024question,
    title = "Question-Based Retrieval using Atomic Units for Enterprise {RAG}",
    author = "Raina, Vatsal  and
      Gales, Mark",
    editor = "Schlichtkrull, Michael  and
      Chen, Yulong  and
      Whitehouse, Chenxi  and
      Deng, Zhenyun  and
      Akhtar, Mubashara  and
      Aly, Rami  and
      Guo, Zhijiang  and
      Christodoulopoulos, Christos  and
      Cocarascu, Oana  and
      Mittal, Arpit  and
      Thorne, James  and
      Vlachos, Andreas",
    booktitle = "Proceedings of the Seventh Fact Extraction and VERification Workshop (FEVER)",
    month = nov,
    year = "2024",
    address = "Miami, Florida, USA",
    publisher = "Association for Computational Linguistics",
    pages = "219--233",
    
}

@article{rasmussen2025zep,
  title={Zep: A Temporal Knowledge Graph Architecture for Agent Memory},
  author={Rasmussen, Preston and Paliychuk, Pavlo and Beauvais, Travis and Ryan, Jack and Chalef, Daniel},
  journal={arXiv preprint arXiv:2501.13956},
  year={2025}
}

@article{zhu2025temporal,
  title={A Temporal Knowledge Graph Generation Dataset Supervised Distantly by Large Language Models},
  author={Zhu, Jun and Fu, Yan and Zhou, Junlin and Chen, Duanbing},
  journal={Scientific Data},
  volume={12},
  number={1},
  pages={734},
  year={2025},
  publisher={Nature Publishing Group UK London}
}

@article{kabal2024enhancing,
  title={Enhancing Domain-Independent Knowledge Graph Construction through OpenIE Cleaning and LLMs Validation},
  author={Kabal, Othmane and Harzallah, Mounira and Guillet, Fabrice and Ichise, Ryutaro},
  journal={Procedia Computer Science},
  volume={246},
  pages={2617--2626},
  year={2024},
  publisher={Elsevier}
}

@inproceedings{yao2019docred,
    title = "{D}oc{RED}: A Large-Scale Document-Level Relation Extraction Dataset",
    author = "Yao, Yuan  and
      Ye, Deming  and
      Li, Peng  and
      Han, Xu  and
      Lin, Yankai  and
      Liu, Zhenghao  and
      Liu, Zhiyuan  and
      Huang, Lixin  and
      Zhou, Jie  and
      Sun, Maosong",
    editor = "Korhonen, Anna  and
      Traum, David  and
      M{\`a}rquez, Llu{\'i}s",
    booktitle = "Proceedings of the 57th Annual Meeting of the Association for Computational Linguistics",
    month = jul,
    year = "2019",
    address = "Florence, Italy",
    publisher = "Association for Computational Linguistics",
    pages = "764--777",

}

@misc{AryanSingh2023NYT,
  author       = {Aryan Singh},
  title        = {{NYT Articles: 2.1M+ (2000-Present) Daily Updated}},
  year         = {2023},
  howpublished = {\url{https://www.kaggle.com/datasets/aryansingh0909/nyt-articles-21m-2000-present}},
  note         = {Accessed: 2025-06-01}
}

@inproceedings{tan2022revisiting,
    title = "Revisiting {D}oc{RED} - Addressing the False Negative Problem in Relation Extraction",
    author = "Tan, Qingyu  and
      Xu, Lu  and
      Bing, Lidong  and
      Ng, Hwee Tou  and
      Aljunied, Sharifah Mahani",
    editor = "Goldberg, Yoav  and
      Kozareva, Zornitsa  and
      Zhang, Yue",
    booktitle = "Proceedings of the 2022 Conference on Empirical Methods in Natural Language Processing",
    month = dec,
    year = "2022",
    address = "Abu Dhabi, United Arab Emirates",
    publisher = "Association for Computational Linguistics",
    pages = "8472--8487",
}

@article{atil2024non,
  title={Non-determinism of" deterministic" llm settings},
  author={Atil, Berk and Aykent, Sarp and Chittams, Alexa and Fu, Lisheng and Passonneau, Rebecca J and Radcliffe, Evan and Rajagopal, Guru Rajan and Sloan, Adam and Tudrej, Tomasz and Ture, Ferhan and others},
  journal={arXiv preprint arXiv:2408.04667},
  year={2024}
}

@article{liu2023lost,
    title = "Lost in the Middle: How Language Models Use Long Contexts",
    author = "Liu, Nelson F.  and
      Lin, Kevin  and
      Hewitt, John  and
      Paranjape, Ashwin  and
      Bevilacqua, Michele  and
      Petroni, Fabio  and
      Liang, Percy",
    journal = "Transactions of the Association for Computational Linguistics",
    volume = "12",
    year = "2024",
    address = "Cambridge, MA",
    publisher = "MIT Press",
    pages = "157--173",
}

@inproceedings{chekol2018towards,
  title={Towards probabilistic bitemporal knowledge graphs},
  author={Chekol, Melisachew Wudage and Stuckenschmidt, Heiner},
  booktitle={Companion Proceedings of the The Web Conference 2018},
  pages={1757--1762},
  year={2018}
}

@article{meijer2022bi,
  title={Bi-VAKs: Bi-temporal Versioning Approach for Knowledge Graphs},
  author={Meijer, Lisa},
  journal={Delft University of Technology},
  year={2022}
}

@article{bian2025llm,
  title={LLM-empowered knowledge graph construction: A survey},
  author={Bian, Haonan},
  journal={arXiv preprint arXiv:2510.20345},
  year={2025}
}

\appendix
\setcounter{figure}{0}
\setcounter{table}{0}
\setcounter{equation}{0}
\setcounter{algorithm}{0}

\renewcommand{\thefigure}{F.\arabic{figure}}    
\renewcommand{\thetable}{T.\arabic{table}}      
\renewcommand{\theequation}{\thesection.\arabic{equation}}
\renewcommand{\thealgorithm}{\thesection.\arabic{algorithm}}

\section{\texttt{ATOM}'s algorithms}
\texttt{ATOM}'s framework is based on two main algorithms, presented below: Algorithm~\ref{alg:binary_atomic_merge} for merging pairs of TKGs and Algorithm~\ref{alg:parallel_atomic_merge} for parallelizing the merge of lists of TKGs. 
\begin{algorithm*}[h]
\footnotesize
\caption{\label{alg:binary_atomic_merge}Binary Merge of TKGs}
\begin{algorithmic}[1]
\Function{BinaryMerge}{$TKG_1 = (\mathcal{E}_1, \mathcal{R}_1)$, $TKG_2 = (\mathcal{E}_2, \mathcal{R}_2)$, $\theta_E$, $\theta_R$}
    \Statex \hspace{\algorithmicindent} \textit{// ------ Entity Resolution ------ //}
    \State \textbf{Initialize} mapping $M \gets \emptyset$
    \ForAll{entity $e \in \mathcal{E}_1$}
        \If{there exists $e' \in \mathcal{E}_2$ such that $e.\text{name} = e'.\text{name}$ and $e.\text{label} = e'.\text{label}$}
            \State $M(e) \gets e'$
        \Else
            \State Compute $s \gets \max\limits_{e' \in \mathcal{E}_2} \cos\Big(e.\mathbf{v}, e'.\mathbf{v}\Big)$  \Comment{Cosine similarity of e and e' embeddings}
            \State Let $e^* \gets \arg\max\limits_{e' \in \mathcal{E}_2} \cos\Big(e.\mathbf{v}, e'.\mathbf{v}\Big)$
            \If{$s \ge \theta_E$}
                \State $M(e) \gets e^*$
            \Else
                \State $M(e) \gets e$
            \EndIf
        \EndIf
    \EndFor
    \State $\mathcal{E}_{\text{merged}} \gets \mathcal{E}_2 \cup \{e \mid M(e) \notin \mathcal{E}_2\}$
    \Statex \hspace{\algorithmicindent} \textit{// ------ Relation's Name Resolution ------ //}
    \State $\mathcal{R}_1^{\text{updated}} \gets \emptyset$
    \ForAll{relation $r \in \mathcal{R}_1$}
        \State \textbf{Update endpoints:} \quad $r.\text{startEntity} \gets M(r.\text{startEntity})$, \quad $r.\text{endEntity} \gets M(r.\text{endEntity})$
        \State Compute $s_r \gets \max\limits_{r' \in \mathcal{R}_2} \cos\Big(r.\mathbf{v}, r'.\mathbf{v}\Big)$ \Comment{Cosine similarity of r and r' names embeddings}
        \State Let $r^* \gets \arg\max\limits_{r' \in \mathcal{R}_2} \cos\Big(r.\mathbf{v}, r'.\mathbf{v}\Big)$
        \If{$s_r \ge \theta_R$}
           
            \State \textbf{Update names:} Update $r.\text{name} \gets r^*.\text{name}$
        \EndIf
        
        \textit{// ------ Temporal Resolution ------ //}
        \If{there exists $r' \in \mathcal{R}_2$ such that $r$ is similar to $r'$} 
         
         \textit{// For similar relations, their times are merged}
            \State \textbf{Update start time:} \quad $r'.t_{start} \gets r'.t_{start} \cup r.t_{start}$
            \State \textbf{Update end time:} \quad $r'.t_{end} \gets r'.t_{end} \cup r.t_{end}$
            \State \textbf{Update observation time:} \quad $r'.t_{obs} \gets r'.t_{obs} \cup r.t_{obs}$
        \EndIf
        \State $\mathcal{R}_1^{\text{updated}} \gets \mathcal{R}_1^{\text{updated}} \cup \{ r \}$
    \EndFor
    \State $\mathcal{R}_{\text{merged}} \gets \mathcal{R}_2 \cup \mathcal{R}_1^{\text{updated}}$
    \State \Return $(\mathcal{E}_{\text{merged}}, \mathcal{R}_{\text{merged}})$
\EndFunction
\end{algorithmic}
\end{algorithm*}
\begin{algorithm*}[h]
\footnotesize  
\caption{Parallel Merge of TKGs}
\label{alg:parallel_atomic_merge}
\begin{algorithmic}[1]
\Function{ParallelMerge}{$TKGs,\, \theta_E,\, \theta_R$}
    \State \textbf{Input:} A list of temporal knowledge graphs 
    \[
      TKGs = \{ TKG_1, TKG_2, \ldots, TKG_n \}
    \]
    \State \(current \gets TKGs\)
    \While{\(|current| > 1\)}
        \State \(mergedResults \gets \emptyset\)
        \State Let \(n \gets |current|\)
        \State \textbf{Form pairs:} 
        \[
          pairs \gets \{\, (current[2i],\, current[2i+1]) \mid 0 \le i < \lfloor n/2 \rfloor \,\}
        \]
        \If{\(n\) is odd}
            \State \(leftover \gets current[n-1]\)
        \Else
            \State \(leftover \gets \text{null}\)
        \EndIf
        \ForAll{each pair \((TKG_a,\, TKG_b)\) in \(pairs\) \textbf{in parallel}}
            \State \(merged \gets \textbf{\textsc{BinaryMerge}}(TKG_a,\, TKG_b,\, \theta_E,\, \theta_R)\)
            \State Add \(merged\) to \(mergedResults\)
        \EndFor
        \If{\(leftover \neq \text{null}\)}
            \State Add \(leftover\) to \(mergedResults\)
        \EndIf
        \State \(current \gets mergedResults\)
    \EndWhile
    \State \Return \(current[0]\)
\EndFunction
\end{algorithmic}
\end{algorithm*}

\section{Example of the atomic fact decomposition}
\label{sec:example-afact-decomp}
Example: \emph{(Observed in 01-01-2025) On June 18, 2024, Real Madrid won the Champions League final with a 2-1 victory. Following the triumph, fans of Real Madrid celebrated the Champions League victory across the city.}
\begin{itemize}
\item \emph{Real Madrid won the Champions League final match on June 18, 2024.} (Observation $t_{obs}=[01-01-2025]$, $t_{\text{start}}=[18-06-2024]$, $t_{\text{end}}=[.]$) 
\item \emph{The Champions League final match ended with a 2-1 victory for Real Madrid on June 18, 2024.} (Observation $t_{obs}=[01-01-2025]$, $t_{\text{start}}=[18-06-2024]$, $t_{\text{end}}=[.]$)
\item \emph{Fans of Real Madrid celebrated the Champions League final match victory across the city on June 18, 2024.} (Observation $t_{obs}=[01-01-2025]$, $t_{\text{start}}=[18-06-2024]$, $t_{\text{end}}=[.]$)
\end{itemize}

\section{Estimating the merging thresholds}
\label{sec:thresholds-estimation}
The merging thresholds $\theta_R$ and $\theta_E$ were estimated by \citep{lairgi2024itext2kg}, based on the mean cosine similarity of 1,500 pairs of similar entities and relation names generated by \texttt{gpt-4-0613}\footnote{\url{https://platform.openai.com/docs/models/gpt-4}}; however, because entity typology is not considered, a hybrid similarity measure is proposed, combining the entity name embedding and the entity label embedding as: $\lambda\,\cdot\,\text{embeddings}_{\text{name}} + \beta\,\cdot\,\text{embeddings}_{\text{label}}$. Using this measure, 1,200 pairs of similar entities incorporating typology were generated, and $\lambda$ is optimized to maximize the resulting cosine similarity, after which it is determined that $\lambda = 0.8$, $\beta = 0.2$, and $\theta_E = 0.8$, while $\theta_R = 0.7$ is retained as previously estimated in \citep{lairgi2024itext2kg}.

\begin{table*}[t]
\centering
\caption{2020-COVID-NYT Statistics Analysis. We use lead paragraphs as they encapsulate the article's key facts, while the full article text is often unavailable in research datasets due to licensing restrictions and would introduce unnecessary verbosity without proportional information gain.}
\label{tab:dataset_statistics}
\begin{tabular}{@{}lc@{}}
\toprule
\textbf{Metric} & \textbf{Value} \\
\midrule
\multicolumn{2}{l}{\textbf{Basic Dataset Information}} \\
\midrule
Total Articles & 1,076 \\
Grouped Articles (by pub. date) & 274 \\
Average Tokens per Group & 206 ± 156 \\
Date Range & 2020-01-09 to 2020-12-30 \\
\midrule
\multicolumn{2}{l}{\textbf{Atomic Facts Analysis}} \\
\midrule
Total atomic facts & 4,223 \\
Atomic facts with validity time & 2,037 \\
Atomic facts without validity time & 2,186 \\
\midrule
\multicolumn{2}{l}{\textbf{Knowledge Graph Structure}} \\
\midrule
Total 5-tuples & 7,210 \\
Number of atomic TKGs & 4,223 \\
Avg number of 5-tuples per atomic TKG & $\sim 2$ \\
\bottomrule
\end{tabular}
\end{table*}

\section{Observation time modeling}
\label{sec:observation-time-modeling}
Modeling the observation time is essential both for capturing the dynamism of the TKG and for inferring relative times. For historical or retrospective unstructured text streams (e.g., past news articles or archive documents), the observation time should correspond to the original publication time rather than the time at which the document was processed and ingested into the DTKG. This distinction is essential for preserving the correct temporal ordering of events and enabling reliable inference of relative times.

Conversely, for prospective or continuously monitored sources, where data is ingested automatically as it becomes available, the observation time can be treated as the ingestion time. In such settings, ingestion reflects the earliest feasible moment at which the information could be known by the system.

The granularity of observation time is application-dependent and may be defined according to user requirements. For instance, COVID-19 news was simulated using daily observation snapshots, whereas social media streams may require a per-post snapshot.

\section{Annotation guidelines for 2020-COVID-NYT}
\label{app:annotation_guidelines}

The ground truth for the 2020-COVID-NYT dataset was constructed through a two-step semi-manual annotation process utilizing the Claude Sonnet 4 chat model with extended thinking\footnote{\url{https://claude.ai/}}. We selected this model after observing its superior performance on a small subset of lead paragraphs compared to other available SOTA LLMs. Human annotators supervised the process, provided with the lead paragraph and an associated observation date ($t_{obs}$), and were instructed to follow the guidelines detailed below.

\subsection{Phase 1: atomic fact decomposition}
In this phase, the model decomposed complex paragraphs into atomic, self-contained, and temporally grounded statements. Annotators were tasked with verifying and correcting the exhaustivity of the decomposition, ensuring the absence of hallucinatory facts, and validating the correctness of validity times for temporal facts, strictly following these guidelines:

\paragraph{Atomicity:} Compound sentences must be split. Each atomic fact must contain exactly one piece of information. Redundancies and duplicated information were removed.
\paragraph{Decontextualization:} Facts must be understandable in isolation. Pronouns (e.g., "he", "it") were replaced with full entity names, and necessary modifiers were included.
\paragraph{Temporal normalization:} All relative time references were converted to absolute dates based on the $t_{obs}$. For example:
\textit{Today} $\rightarrow$ exact $t_{obs}$; \quad \textit{Yesterday} $\rightarrow$ $t_{obs} - 1$ day; \quad \textit{Last week} $\rightarrow$ Monday of the week preceding $t_{obs}$; \quad \textit{This year} $\rightarrow$ January 1st of the $t_{obs}$ year.

\paragraph{End actions:} If the text indicated the end of a role or action (e.g., "leaving a position"), the termination was captured along with its specific timestamp.

\subsection{Phase 2: 5-tuple extraction}
In the second phase, the model extracted temporal 5-tuples $(e_s, r, e_o, t_{\text{start}}, t_{\text{end}})$ from the atomic facts generated in Phase 1. Similarly, annotators verified and corrected the extracted tuples for exhaustivity, factual accuracy, and strict temporal validity, ensuring adherence to the following guidelines:

\paragraph{Entity Annotation}
Entities were defined as distinct concepts (e.g., Person, Organization, Position). Dates were strictly excluded from entity mentions (they are reserved for temporal arguments).

\paragraph{Relationship and temporal annotation}
Relations were extracted to capture the interaction between entities, with strict temporal rules:
\begin{itemize}
    \item \textbf{Canonical present tense:} All relation names were annotated in the present tense (e.g., \textit{is\_CEO}, \textit{works\_at}) regardless of whether the event was past, present, or future. This enhances semantic consistency across the TKG.
    \item \textbf{Validity periods:} Temporal information was mapped strictly to $t_{\text{start}}$ (validity start) and $t_{\text{end}}$ (validity end) based on the absolute dates resolved in Phase 1.
    \begin{itemize}
        \item \textit{Affirmative actions} (e.g., "became CEO") populated $t_{\text{start}}$.
        \item \textit{End actions} (e.g., "is no longer the CEO") populated $t_{\text{end}}$, while the relation name remained in the affirmative (e.g., \textit{is\_CEO}).
    \end{itemize}
\end{itemize}

\begin{figure*}[h]
  \centering
  \includegraphics[width=\textwidth]{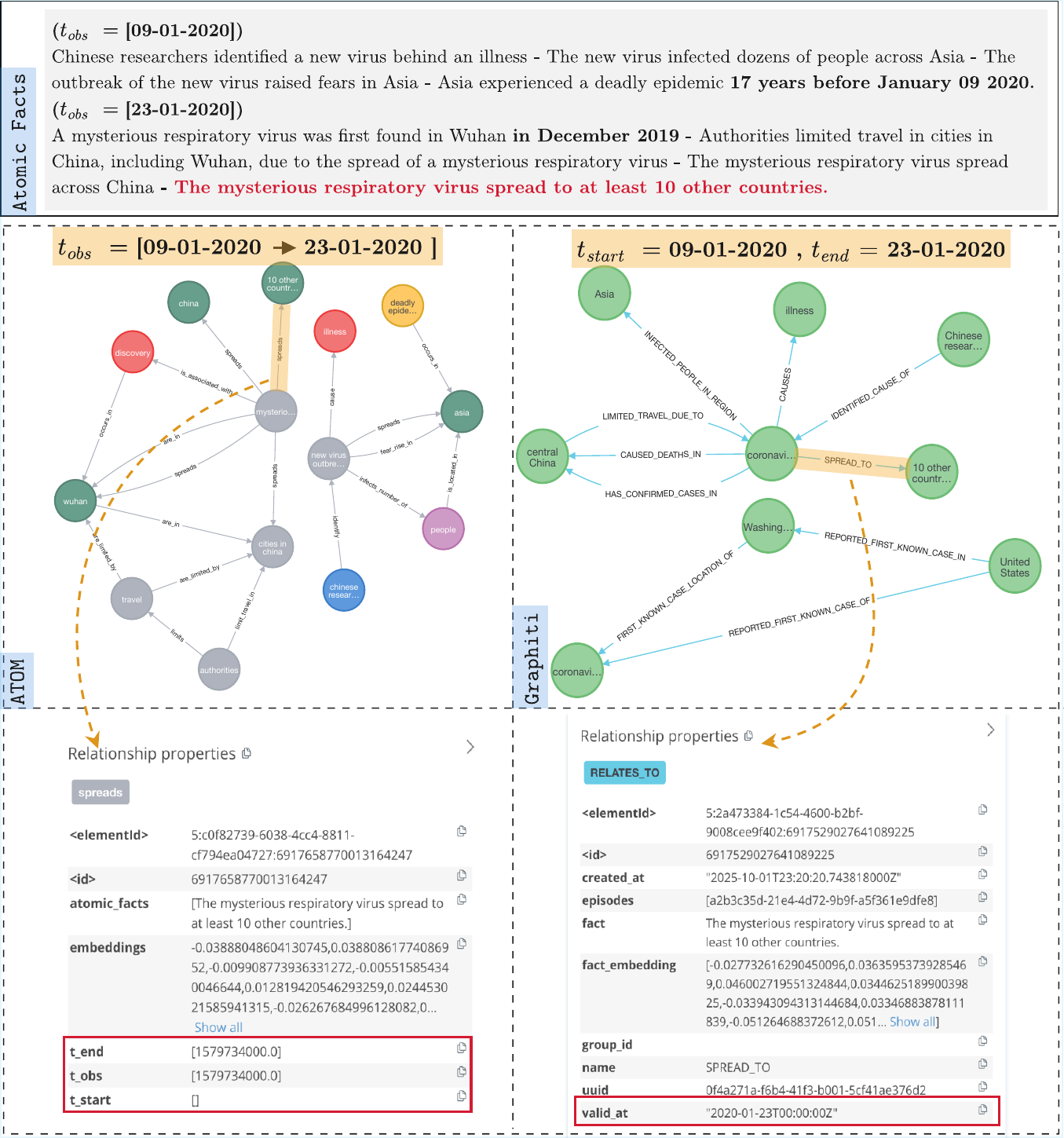}
  \caption{{ Two DTKGs constructed using \texttt{ATOM} and Graphiti from 09-01-2020 (in UNIX, 1578524400) to 23-01-2020 (in UNIX, 1579734000) from 2020-COVID-NYT dataset. \textbf{Left (\texttt{ATOM}):} Preserves observation times ($t_{obs}$) separately from validity periods, with timestamps encoded in UNIX format to eliminate overhead associated with string parsing operations and timezone conversion calculations. \textbf{Right (Graphiti):} Treats observation time as validity start time. $valid\_at$ corresponds to $t_{\text{start}}$ in Graphiti's time modeling. The highlighted fact \textit{``The mysterious respiratory virus spread to at least 10 other countries''} is observed on 23-01-2020, but this does not guarantee the spread occurred at that time. \texttt{ATOM}'s dual-time modeling prevents such temporal misattribution.}
}
  \label{fig:atomvsgraphiti}
\end{figure*}

\begin{figure*}[h]
  \centering
  \includegraphics[width=\textwidth]{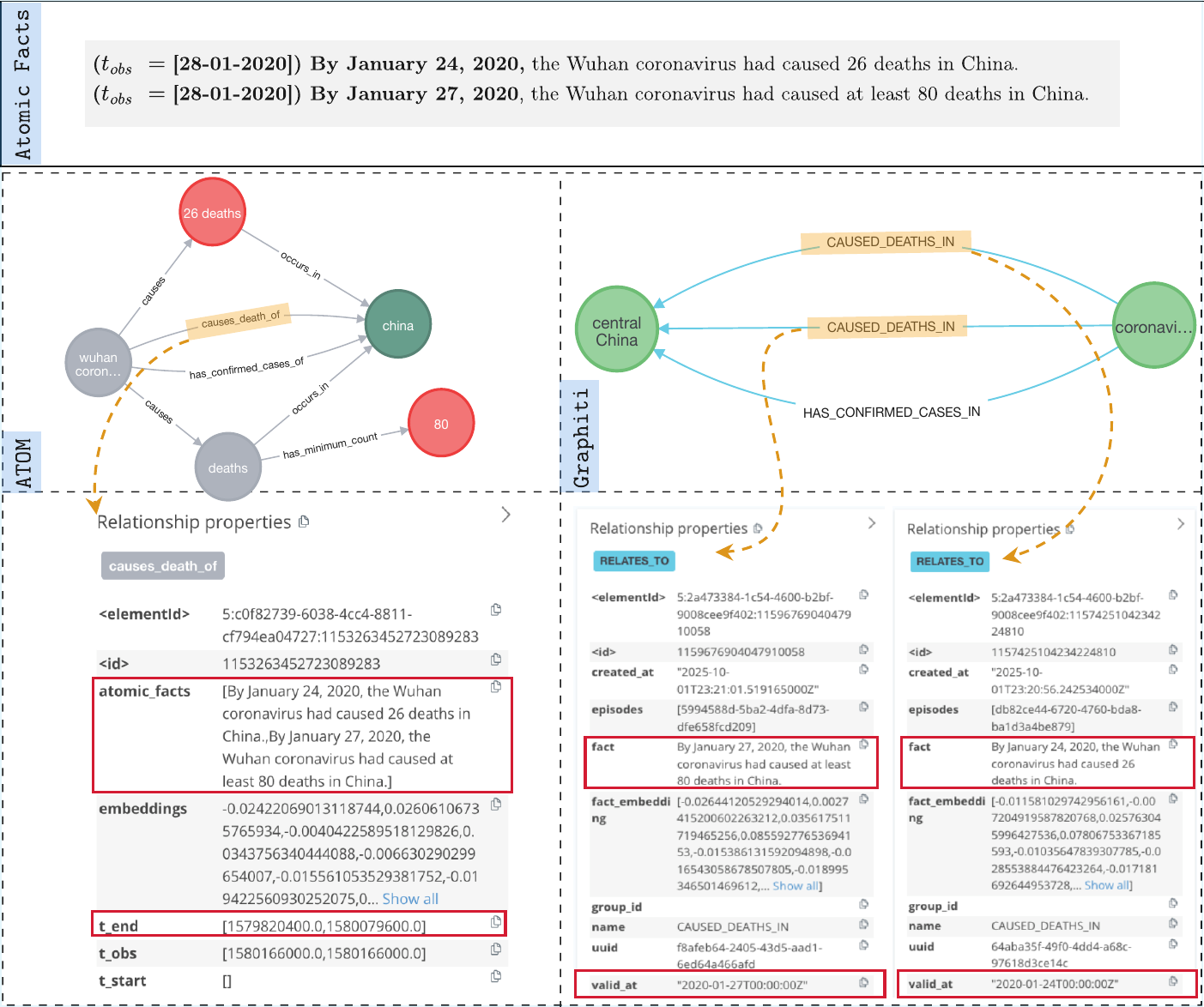}
  \caption{{Temporal resolution comparison between \texttt{ATOM} and Graphiti. Two atomic facts observed on January 28, 2020, report death counts from January 24 (26 deaths) and January 27 (at least 80 deaths). \textbf{Left (\texttt{ATOM})}: performs temporal resolution by detecting similar relations and extending their validity period history ($t_{\text{end}}$ in the figure). \textbf{Right (Graphiti)}: creates separate relations for each atomic fact, resulting in duplication. Moreover, Graphiti misinterprets \textit{``By January 24, 2020''} and \textit{``By January 27, 2020''} as validity start times rather than validity end times, leading to temporal misattribution.}
}
  \label{fig:atomvsgraphiti_death}
\end{figure*}

\begin{figure*}[h]
  \centering
  \includegraphics[width=\textwidth]{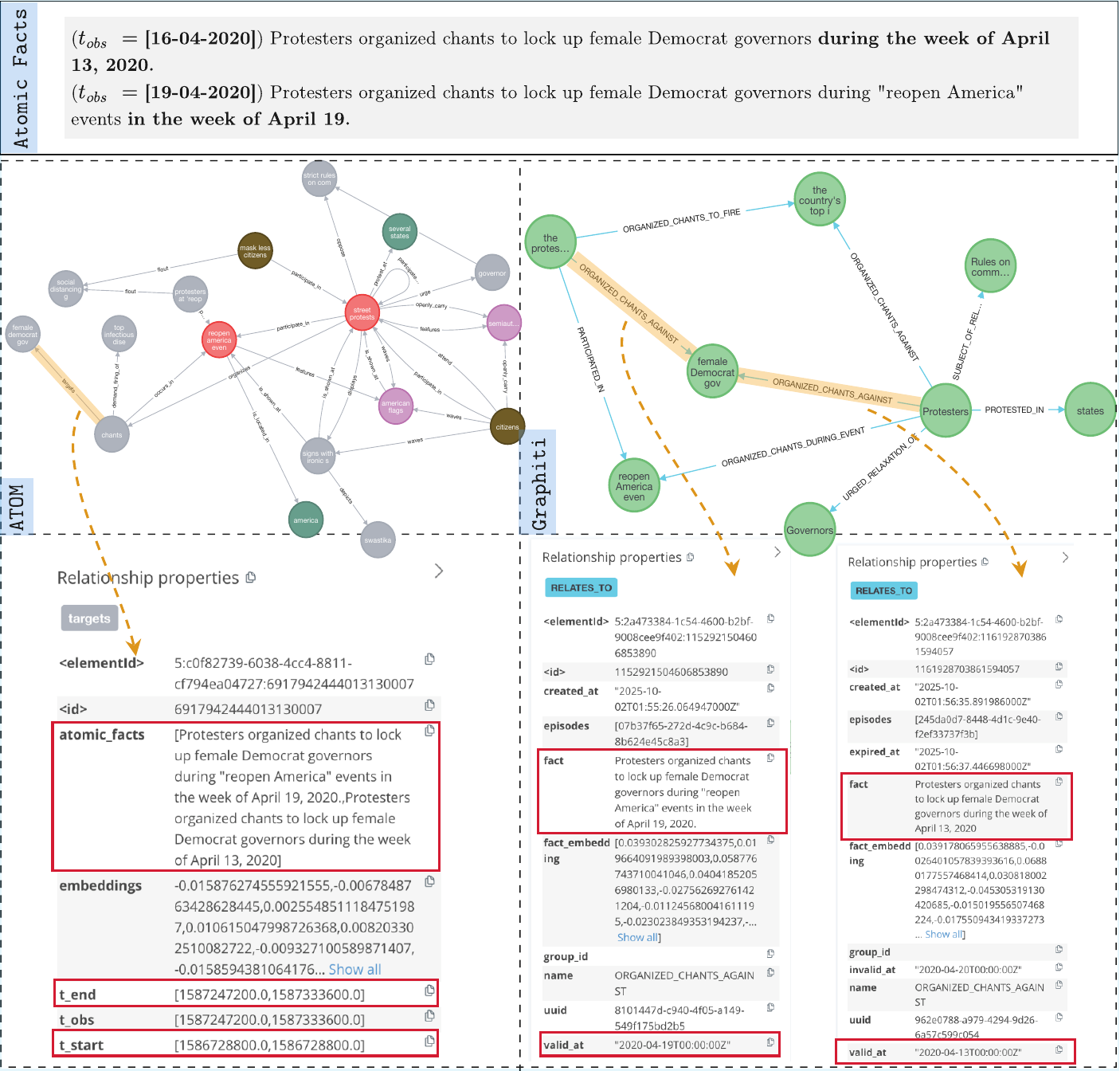}
  \caption{{Temporal resolution comparison between \texttt{ATOM} and Graphiti. Two atomic facts observed on different dates (April 16 and April 19, 2020) describe protest activities during two time periods (\textit{``the week of April 13''} and \textit{``the week of April 19''}). \textbf{Left (\texttt{ATOM})}: merges similar relations and extend their validity periods ($t_{\text{start}}$ and $t_{\text{end}}$ in the figure). \textbf{Right (Graphiti)}: maintains separate relations for each atomic fact. Moreover, Graphiti fails to translate \textit{``In the week of April 13, 2020''} and \textit{``In the week of April 19, 2020''} into proper validity periods as \texttt{ATOM} does.}}
  \label{fig:atomvsgraphiti_protests}
\end{figure*}

\begin{table*}[t]
\centering
\caption{Cost and token usage analysis on the 2020-COVID-NYT dataset. \textbf{(L)} denotes direct extraction from Lead paragraphs. \textbf{(F)} denotes the full \texttt{ATOM} pipeline, broken down into \textbf{Module-1} (atomic fact decomposition) and \textbf{Module-2} (5-tuple extraction from atomic facts).}
\label{tab:cost_estimation_detailed}
\begin{tabular}{@{}lcccc@{}}
\toprule
\textbf{} & \textbf{Direct Extraction} & \multicolumn{3}{c}{\textbf{ATOM (F)}} \\
\cmidrule(l){3-5}
\textbf{Metric} & \textbf{(L)} & \textbf{Module-1} & \textbf{Module-2} & \textbf{Total} \\
\midrule
\multicolumn{5}{l}{\textbf{Token Usage Statistics}} \\
\midrule
Total Input Tokens      & 271,108 & 237,192 & 285,618 & 522,810 \\
Total Output Tokens     & 83,757  & 38,646  & 217,110 & 255,756 \\
Avg Input per Article   & 2,222   & 1,944   & 2,341   & 4,285   \\
Avg Output per Article  & 687     & 317     & 1,780   & 2,096   \\
\midrule
\multicolumn{5}{l}{\textbf{Estimated Cost (USD)}} \\
\midrule
\texttt{claude-sonnet-4-2025-01-31}         & \$1.04  & \$0.64  & \$2.06  & \$2.70  \\
\texttt{gpt-4.1-2025-04-14}                 & \$0.61  & \$0.40  & \$1.15  & \$1.55  \\
\bottomrule
\end{tabular}
\end{table*}

\end{document}